\documentclass[conference]{IEEEtran}
\usepackage[utf8]{inputenc}
\usepackage[T1]{fontenc}
\usepackage{amsmath,amssymb}
\usepackage{graphicx}
\usepackage{booktabs}
\usepackage{multirow}
\usepackage{xcolor}
\usepackage{url}
\usepackage[hidelinks]{hyperref}
\usepackage{listings}
\usepackage{float}
\usepackage{balance}
\begin{document}

\title{The LLM Proposes, the Benchmark Disposes:\\Center Bias in LLM-Driven Algorithm Design}

\author{\IEEEauthorblockN{Yuefeng Xu\IEEEauthorrefmark{1}, Rui Zhong\IEEEauthorrefmark{2}, Jun Yu\IEEEauthorrefmark{3}}
\IEEEauthorblockA{\IEEEauthorrefmark{1}Graduate School of Science and Technology, Niigata University, Niigata, Japan\\
f25l005c@mail.cc.niigata-u.ac.jp}
\IEEEauthorblockA{\IEEEauthorrefmark{2}Information Initiative Center, Hokkaido University, Sapporo, Japan\\
zhongrui@iic.hokudai.ac.jp}
\IEEEauthorblockA{\IEEEauthorrefmark{3}Institute of Science and Technology, Niigata University, Niigata, Japan\\
yujun@ie.niigata-u.ac.jp}}

\maketitle

\begin{abstract}
Large language models (LLMs) are increasingly used to design black-box optimization algorithms automatically
(LLM-driven automated algorithm design, LLM-AAD). These designs are selected by their scores on training benchmarks,
and many classic benchmarks place the optimum at or near the center of the search box.
We ask whether LLM-AAD produces optimizers with a center or origin bias that exploits this.
Using the shift-ratio test of K\r{u}dela and the BIAS toolbox, we screen 329 published LLM-generated algorithms from six
sources and, under the same protocol, 330 human-designed metaheuristics. The shift-ratio test flags
1.9\% (6/312) of the LLM-generated algorithms at $D=5$ and 1.8\% (5/276) at $D=10$, against 45.8\% (149/325) and
47.5\% (154/324) of the human-designed ones.
A controlled LLaMEA experiment varies only the training benchmark. With unshifted training functions, 5 of 20 final
algorithms are flagged at $D=10$, versus 0 of 20 with randomly shifted ones (exploratory; Holm-corrected $p=0.094$).
On independent test instances, center-trained algorithms lose more performance when the optimum is moved
(median AOCC gap 0.024 vs.\ 0.005; exploratory one-sided Mann--Whitney $p=0.003$).
Biased code reached the top of the population under both training sets (6/20 vs.\ 3/20 runs) but was retained to the end
more often under unshifted training (5/6 vs.\ 0/3).
A single code change removes the flag for each of the ten biased algorithms we examined. To our knowledge, two mechanisms
have not been reported for LLM-generated optimizers: a velocity clamp equal to the box half-width, which lands particles
exactly on the center, and bugs that leave individuals at the origin. For three of five center-trained algorithms the change
costs no detectable performance on shifted instances, for one a borderline loss, and for a CMA-ES started at the box center
a large one. In our experiment, randomly shifting the training functions kept flagged algorithms from surviving to the end.
\end{abstract}


% =====================================================================
\section{Introduction}
LLM-driven automated algorithm design (LLM-AAD) uses a large language model as the variation operator of an
evolutionary search over programs: the LLM writes or mutates an optimizer, the optimizer is run on a training
benchmark, and its score decides what survives. Frameworks such as LLaMEA~\cite{vanstein2025llamea},
EoH~\cite{liu2024eoh} and ReEvo~\cite{ye2024reevo} have produced competitive continuous optimizers this way;
LLMs have also been prompted to invent new metaheuristics directly~\cite{zhong2024zso};
and LLM-generated algorithms now enter competitions and benchmark suites~\cite{vanstein2025blade}.

The score is the only thing the search sees. If the training benchmark has a regularity that a program can exploit,
selection will reward the exploit. A well-known regularity of continuous benchmarks is the location of the optimum:
many classic test functions have it at the origin, which is also the center of the usual symmetric search box.
For human-designed metaheuristics, K\r{u}dela showed that such a \emph{center bias} is common and can make
algorithms look far better than they are: about half of 90 recently proposed methods deteriorate sharply once the
optimum is moved away from the center~\cite{kudela2022critical,kudela2023noone}.
For LLM-AAD the question is open. A single generated Bayesian-optimization algorithm was observed to ``cheat'' on
BBOB $f_9$ and $f_{19}$ by sampling near the origin~\cite{llameabo2025}, and recent best-practice guidelines list
invariance tests as rarely performed~\cite{vanstein2026bestpractices}, but no study has measured how common the
problem is, where it comes from, or which code causes it.

We address three questions:
\begin{itemize}
\item \textbf{Prevalence.} How many published LLM-generated optimizers have a center or origin bias, and how does
  this compare with human-designed metaheuristics?
\item \textbf{Source.} Is the bias caused by the training benchmark (selection rewards a shortcut), or by the LLM's prior
  (biased code learned from existing metaheuristics), or both?
\item \textbf{Mechanism.} Which code is responsible, and does removing it cost real performance?
\end{itemize}

Our contributions are:
(i) a same-protocol screening of 329 published LLM-generated algorithms and 330 human-designed metaheuristics, with two
complementary detectors validated on eleven controls (Section~\ref{sec:prevalence});
(ii) a controlled LLaMEA experiment that differs only in the location of the training optimum, with an evaluation on
independent test instances and a generation-by-generation timeline of when biased code appears and whether it is retained
(Section~\ref{sec:source});
(iii) single code changes that remove the bias flag for each of the ten biased algorithms examined, identifying the responsible
code, including mechanisms that, to our knowledge, have not been reported for LLM-generated optimizers
(Section~\ref{sec:mechanism}); and
(iv) practical recommendations for LLM-AAD benchmarks (Section~\ref{sec:discussion}).

% =====================================================================
\section{Background and Related Work}
\label{sec:related}
\textbf{Center and structural bias.}
An algorithm has a \emph{center bias} if it tends to sample or converge near the center of the search box regardless of
the objective, and an \emph{origin bias} if its operators implicitly use the coordinate origin, e.g.\ a multiplicative
contraction $x \leftarrow c\,x$ toward~0. On a symmetric box $[-B,B]^D$ the two coincide; an asymmetric box separates them.
An operator of the form $\sum_i w_i x_i + \sum_k c_k (x_a - x_b)$ is translation invariant if and only if $\sum_i w_i = 1$
\cite{tian2021invariant}. K\r{u}dela's shift-ratio test~\cite{kudela2022critical,kudela2023noone} compares an algorithm's
error on a function with the optimum in place and shifted by 10\% of the range; a geometric-mean ratio above 10 over a
test suite indicates center bias. The BIAS toolbox~\cite{vermetten2022bias,kononova2015structural} instead runs an
algorithm on a random objective $f_0$ and tests whether its final points are uniformly distributed;
Deep-BIAS~\cite{vanstein2023deepbias} classifies the bias type.

\textbf{LLM-driven algorithm design.}
LLaMEA~\cite{vanstein2025llamea} evolves Python optimizer classes with an LLM inside an evolution strategy and scores
them by the area over the convergence curve (AOCC) on BBOB-type functions~\cite{hansen2021coco}; BLADE~\cite{vanstein2025blade}
collects such runs across benchmark suites; LLaMEA-BO~\cite{llameabo2025} targets Bayesian optimization.
The closest prior observation is Appendix~D of~\cite{llameabo2025}: one generated algorithm exploited the near-center
optimum of BBOB $f_9$/$f_{19}$, and those instances were excluded.
Benchmarking guidelines for LLM-AAD~\cite{vanstein2026bestpractices} rate invariance testing, including invariance to
translation, at the lowest adoption level (L0); they also caution that the generation process may itself bias generated
algorithms toward the representations seen during search, and note that no study has tested this explicitly.
They do not discuss center or origin bias, and a behavior-space analysis of LLaMEA runs~\cite{vanstein2026behaviour} characterizes
exploration and exploitation without testing translation invariance. BBOB itself places the optimum of $f_9$, $f_{19}$,
$f_{20}$ and $f_{24}$ differently from the other functions~\cite{long2022bbobinstances}.

% =====================================================================
\begin{figure*}[t]
\centering
\includegraphics[width=\textwidth]{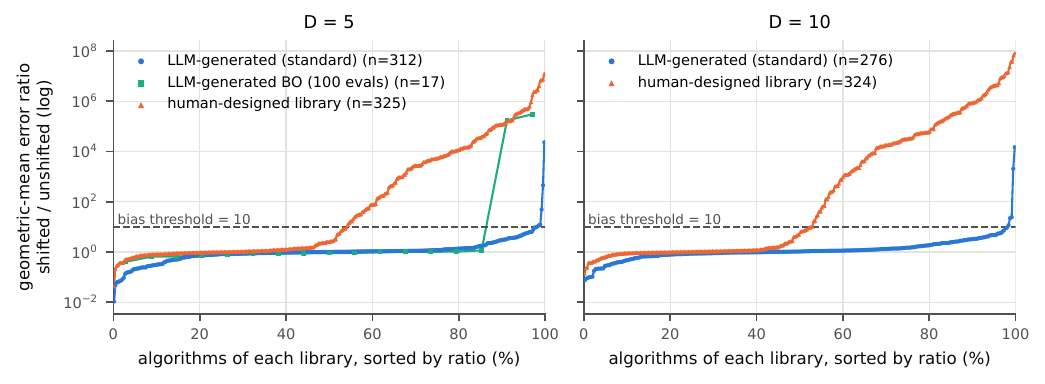}
\caption{Shift ratio of every screened algorithm (log scale), each library sorted separately; the $x$-axis is the rank as a
percentage of the library. The dashed line is K\r{u}dela's threshold of 10.}
\label{fig:ratios}
\end{figure*}

\section{Methods}
\label{sec:methods}
\subsection{Detectors}
\textbf{Shift-ratio test.} We follow K\r{u}dela~\cite{kudela2023noone} with the 13 functions F01--F13 of his Table~1
(Sphere to Penalized~2; optimum at the origin for eight of them). Each function is run unshifted and shifted,
$f(\mathbf{x}+\mathbf{s})$ with $s_j$ equal to 10\% of the range, so every coordinate of the optimum moves by $-s_j$
(by $-1$ in the rescaled coordinates below). The per-run error $f-f^*$ is floored at $10^{-8}$, averaged
over runs, and the ratio shifted/unshifted is computed per function; the geometric mean over functions above 10 means
\emph{biased}. Because the LLM-generated algorithms were designed for $[-5,5]^D$, we express every function in rescaled
coordinates $y=5x/B$ so that the box is $[-5,5]^D$ (function values unchanged), and use $D\in\{5,10\}$, a budget of
$2000D$ evaluations and 30 runs (K\r{u}dela: $D=30$, $5\times10^4$ evaluations, 20 runs; Supplementary Material~S1). LLaMEA-BO algorithms, designed
for 100 evaluations, use $D=5$ and 100 evaluations. Failed runs (exception, time limit, no in-bounds evaluation) are excluded from the mean error, and a verdict is valid if
every function has at least 10 successful runs in both versions; recomputing the valid verdicts with failed runs set to the
worst error of the function changes none of them. We add a third version with the asymmetric box $[-B/2,3B/2]^D$ (optimum unchanged) to separate origin from
center bias; it is only interpretable when an algorithm respects the given bounds. ``Biased'' and ``flagged'' below
always mean a valid geometric-mean ratio above 10 under this protocol, at the stated dimension.

\textbf{BIAS.} Each algorithm is run 100 times on $f_0\sim U(0,1)$ in $[-5,5]^D$ with budget $10^4D$ for $D\in\{5,30\}$;
final points are rescaled to $[0,1]$ and passed to the BIAS test battery and Deep-BIAS~\cite{vermetten2022bias,vanstein2023deepbias}.
BIAS verdicts exist for 310 and 275 of the 312 standard-protocol LLM-generated algorithms: at $D=5$ two algorithms failed all
BIAS runs, and at $D=30$ the 36 algorithms that hard-code $D=5$ cannot run and one more completed only 14 of 100 runs.

\textbf{Harness.} All algorithms receive the objective through a wrapper that counts evaluations, stops the run at the
budget, and records the best \emph{in-bounds} point actually evaluated, so an algorithm cannot misreport its result.
Runs are isolated per process with a hard time limit (900\,s). Eleven controls with known behavior under this test (GWO, WOA, HHO, SCA and SMA reported biased by K\r{u}dela;
DE, PSO, CMA-ES, random search, MFO and ALO not flagged by the shift-ratio test) are classified as expected at both
dimensions (11/11).

\subsection{Algorithm libraries}
\textbf{LLM-generated.} We collected 333 algorithms (330 distinct) from six public sources: BLADE runs on SBOX-COST (199) and
MA-BBOB (50)~\cite{vanstein2025blade}, the LLaMEA paper's runs (45) and its EoH baseline (5)~\cite{vanstein2025llamea},
LLaMEA-BO (20)~\cite{llameabo2025}, LLaMEA-HPO (8), the LLaMEA repository (2), and four entries of the GECCO 2025/2026
LLM-designed-EA competitions. One competition entry (MSHO-LLM) calls an LLM at run time and is not screened, leaving 329
distinct algorithms (312 for the standard protocol and 17 Bayesian-optimization algorithms for the low-budget protocol).
Most (319) are the best algorithm of a run, not a hand-picked showcase.
They were generated by at least eight LLMs (Gemini 1.5/2.0 Flash, GPT-3.5/4/4o, Codestral, Qwen2.5-Coder and others).

\textbf{Human-designed.} To compare at the same scale, we screen 330 human-designed metaheuristics: all 231 usable
optimizer classes of mealpy 3.0.3~\cite{vanthieu2023mealpy} (136 base algorithms and published variants; one class is excluded
because of a library bug) and 99 algorithms of pyMetaheuristic 7.4.4~\cite{pereira2026pymetaheuristic}, drawn with a fixed seed
from the 260 engines that ran, cite a DOI and are not in mealpy. All use library defaults and exactly the protocol above
(772{,}200 runs). The eleven controls (GWO, WOA, HHO, SCA, SMA, MFO and ALO from mealpy; DE, PSO, CMA-ES and random
search) are run under the same protocol.

\subsection{Controlled evolution (experiment B)}
We ran LLaMEA 1.4.2 with its default $(5{+}5)$ strategy and 200 LLM-generated candidates per run, using
Qwen2.5-Coder-14B~\cite{hui2024qwen25coder} (local, temperature 0.8). The training set is F01--F13 without F08
(whose optimum lies near the boundary), $D=5$, budget $10^4$, two runs per function; the score is the mean AOCC over these runs. For one run on function $f$,
$q(e)=\min\{1,\max\{0,(u_f-\log_{10}\max(e,10^{-8}))/(u_f+8)\}\}$, where $e$ is the best-so-far error over
in-bounds evaluations and $u_f$ the $\log_{10}$ of the median error of 200 uniform random points of $f$; the AOCC is the mean
of $q$ over the budget, with $q=0$ before the first in-bounds evaluation and the last value carried to the end if the
algorithm stops early. The two groups differ \emph{only} in the training functions: \emph{centered} uses them unshifted, so the optimum lies at
or near the origin (exactly at it for 8 of the 12 functions; within 0.17 of it per coordinate for Rosenbrock, Step and the two
Penalized functions, in $[-5,5]$ coordinates); \emph{shifted} draws a new uniform random shift for every optimization run of
every candidate: the function is translated so that its optimum (also for the four functions whose unshifted optimum is
not at the origin) lies at a uniform random point of $[-4,4]^D$, and the shift stays fixed during that run.
Prompt and feedback are identical, and both state the box $[-5,5]^D$ (Supplementary Material~S4). Each group has 20 independent runs.
Every best-so-far algorithm (497 in total) is screened with the shift-ratio test at $D=5$ and $D=10$ with 10 runs
instead of 30, and counted as biased if a valid ratio exceeds 10 at either dimension; final algorithms are screened with the
full protocol and reported per dimension. Each run's final algorithm is then evaluated on unseen instances of the training functions: all algorithms run on
the same pre-generated instances, unshifted and with one random shift per function and run (fixed during the run), five runs
each, at $D=5$ as in training, with optimizer and noise seeds disjoint from those used in training. The ablations of
Section~\ref{sec:mechanism} use the same instances.

\subsection{Statistics}
Proportions carry Wilson 95\% intervals; $2\times2$ tables use Fisher's exact test with a Haldane-corrected odds ratio (OR);
two-sample comparisons use the Mann--Whitney $U$ test with the rank-biserial correlation $r$. Only one comparison was
specified in our experiment plan: the two-sided Fisher test of flagged final algorithms between the groups, at $D=5$ and
$D=10$, with Holm's correction over these two tests. All other tests of the controlled experiment (ratio sizes, independent
test scores, timeline), including the direction of the one-sided ones, were chosen after the data were seen; we report
them uncorrected as exploratory. Fisher tests treat algorithms as independent, although many share code or ancestry.

% =====================================================================
\begin{figure*}[t]
\centering
\includegraphics[width=0.9\textwidth]{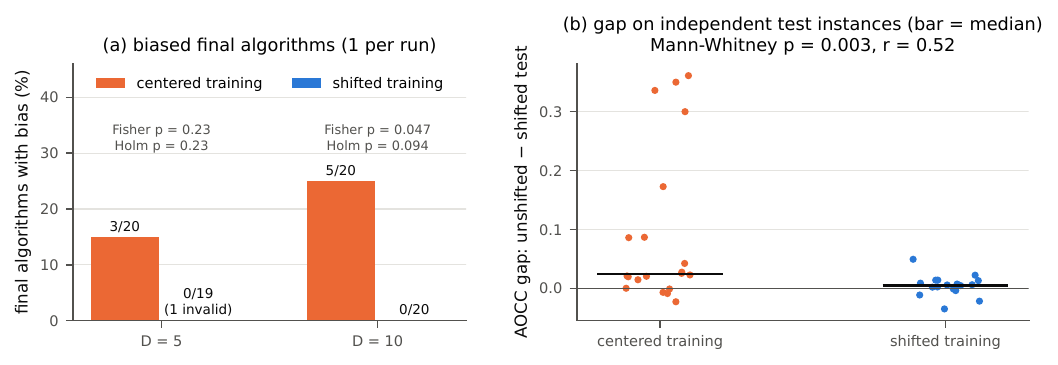}
\caption{Controlled evolution, 20 runs per group. (a) Final algorithms flagged as biased (one shifted-group verdict at $D=5$
is invalid). (b) AOCC gap of each final algorithm between unshifted and shifted instances, on independent test instances
shared by all algorithms; bars are medians.}
\label{fig:groups}
\end{figure*}

\begin{figure}[t]
\centering
\includegraphics[width=\columnwidth]{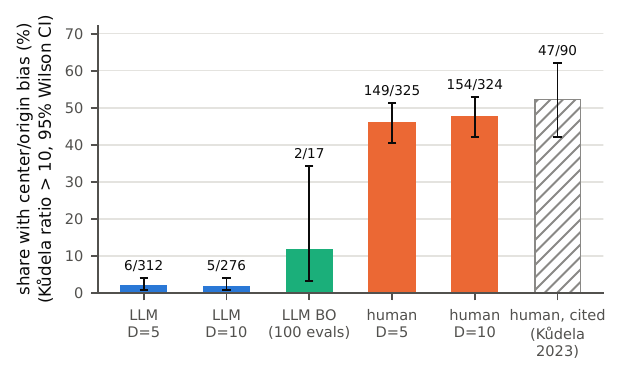}
\caption{Share of algorithms with center/origin bias, with Wilson 95\% intervals. Solid bars: our protocol. Hatched: value
cited from K\r{u}dela~\cite{kudela2023noone} under a different protocol.}
\label{fig:prevalence}
\end{figure}

\begin{table}[t]
\centering
\caption{Flagged LLM-generated algorithms by source (valid verdicts). 36 algorithms of the LLaMEA paper hard-code $D=5$.}
\label{tab:sources}
\begin{tabular}{lrr}
\toprule
Source & $D=5$ & $D=10$\\
\midrule
BLADE, SBOX-COST runs & 4/199 & 2/199\\
BLADE, MA-BBOB runs & 1/50 & 2/50\\
LLaMEA paper runs & 1/45 & 1/9\\
LLaMEA paper, EoH baseline & 0/5 & 0/5\\
LLaMEA-HPO & 0/8 & 0/8\\
LLaMEA repository & 0/2 & 0/2\\
GECCO competitions & 0/3 & 0/3\\
\midrule
All, standard protocol & 6/312 & 5/276\\
LLaMEA-BO (100 evaluations) & 2/17 & --\\
\bottomrule
\end{tabular}
\end{table}

\section{Prevalence}
\label{sec:prevalence}

Fig.~\ref{fig:ratios} shows that almost all LLM-generated algorithms have shift ratios close to 1.
Six of 312 valid verdicts exceed the threshold at $D=5$ (1.9\%, 95\% CI 0.9--4.1\%) and five of 276 at $D=10$
(1.8\%, 0.8--4.2\%); 36 algorithms hard-code $D=5$ and cannot be tested at $D=10$.
Low-budget Bayesian-optimization algorithms are more often flagged (2/17, 11.8\%, 3.3--34.3\%).
The low rate is not driven by one source (Table~\ref{tab:sources}): BLADE contributes 249 of the 312 algorithms, and the
other sources together give 1/63 at $D=5$ and 1/27 at $D=10$.
Under the same protocol, the five known-biased human-designed controls reach ratios of 919, 15, 114{,}251, 8{,}014 and
13{,}037 at $D=5$ (GWO, WOA, HHO, SCA, SMA; 61--40{,}212 at $D=10$), the same range as the few biased LLM-generated
algorithms: when bias occurs it is as strong in LLM-generated code; what differs is how often it occurs.

\textbf{Human-designed library.} Under the identical protocol, 149 of 325 valid human-designed algorithms are biased at
$D=5$ (45.8\%, 95\% CI 40.5--51.3\%) and 154 of 324 at $D=10$ (47.5\%, 42.2--53.0\%), against 1.9\% and 1.8\% of the
LLM-generated ones (Fisher $p=1.6\times10^{-44}$ and $1.1\times10^{-43}$; OR 40 and 45). The bias is strong: the median
ratio of the biased human-designed algorithms is $7.0\times10^{3}$ at $D=5$ and $3.7\times10^{4}$ at $D=10$
(Fig.~\ref{fig:ratios}), and 146 algorithms are biased at both dimensions. The asymmetric box classifies most of them as
origin-biased (128 of 149 at $D=5$, 131 of 154 at $D=10$), as expected for multiplicative operators such as GWO's
$C\cdot X$; 18 and 19 are center-biased. Both libraries agree (mealpy 47.6\%/50.0\%, pyMetaheuristic 41.8\%/41.8\% at
$D=5$/$10$), and the rate is close to the 52\% (47/90) that K\r{u}dela reported under his original protocol ($D=30$, native
bounds; Fig.~\ref{fig:prevalence}, hatched). Implementations matter: the same algorithm can behave very differently in the
two libraries (GWO reaches an error of $10^{-3}$ in mealpy and $6.5\times10^{3}$ in pyMetaheuristic on the shifted Sphere
function at $D=5$ in a test run). Five and six algorithms give invalid verdicts at $D=5$ and $D=10$ (time limit or library
errors). Many mealpy classes are variants of the same base algorithm; counting each base algorithm once
(233 base algorithms with a valid verdict, each weighted equally) gives 46.4\% and 47.2\%, and the variants of a base
algorithm agree in all but 13 of them at $D=5$. The two collections also differ in how they were selected: the LLM-generated
algorithms are mostly the best of a benchmark-driven search, the human-designed ones library defaults. The comparison
therefore describes these collections, not LLM and human design in general.

BIAS flags most LLM-generated algorithms as structurally biased (290/310 at $D=5$, 256/275 at $D=30$), but mostly with
\emph{boundary} bias: about 70\% of the flagged algorithms place more than 5\% of coordinates exactly on the bound, the
signature of clipping, and the most frequent Deep-BIAS class is ``bounds''. Only 19 show samples concentrated in the middle of
the box. BIAS also flags PSO, MFO and ALO, which the shift-ratio test does not flag. The two detectors therefore measure different things:
BIAS measures structural bias on a random objective, the shift ratio measures whether a bias affects performance.
We use the shift ratio for center/origin bias.

% =====================================================================
\section{Source: Selection Keeps a Shortcut that the LLM Proposes}
\label{sec:source}

\begin{figure*}[t]
\centering
\includegraphics[width=\textwidth]{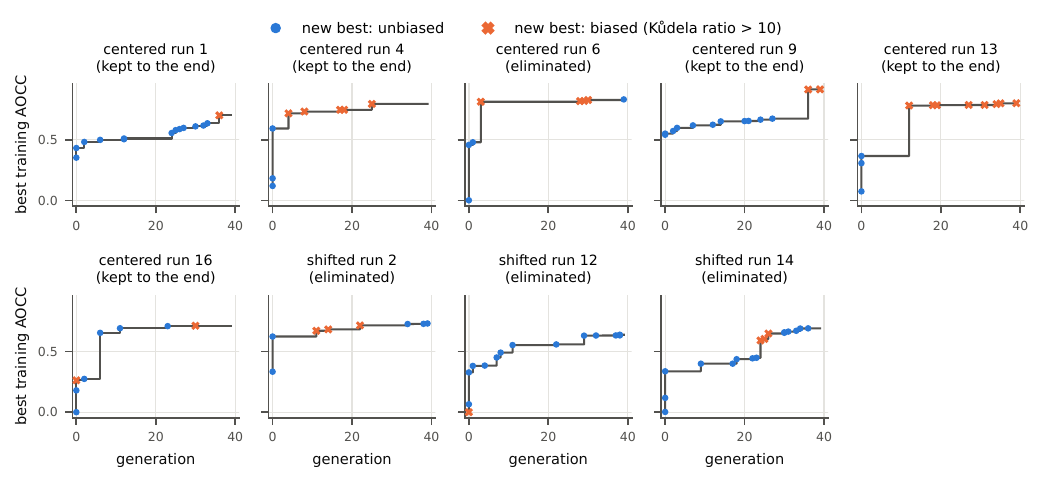}
\caption{Best training AOCC over generations for the nine runs in which a biased best-so-far algorithm appeared.
Crosses mark new best algorithms with shift ratio above 10.}
\label{fig:timeline}
\end{figure*}

\textbf{Final algorithms.} No final algorithm of the shifted group was found biased (0/19 valid verdicts at $D=5$, one verdict being invalid; 0/20 at
$D=10$), whereas 3 of 20 (at $D=5$) and 5 of 20 (at $D=10$) of the centered group were (Fig.~\ref{fig:groups}a). Two-sided Fisher tests, the comparison specified in our experiment plan, give $p=0.23$ (OR 7.8) and $p=0.047$ (OR 14.6);
after Holm correction over the two dimensions, $p=0.23$ and $p=0.094$, so this comparison alone is not conclusive.
The sample size was not fixed in advance: after five runs per group (Fisher $p=1.0$ and $0.44$) we extended both groups to
20 runs. These results are therefore exploratory evidence, not a confirmatory test. The ratios themselves are larger in the centered group
(exploratory one-sided Mann--Whitney: $D=5$, $U=288$, $p=0.003$, $r=0.52$; $D=10$, $U=255$, $p=0.070$, $r=0.28$).

\textbf{Performance when the optimum moves.} On independent test instances of the training functions
(Fig.~\ref{fig:groups}b), center-trained algorithms lose more AOCC between the unshifted and the shifted instances than
shift-trained ones: median 0.024 vs.\ 0.005, mean 0.092 vs.\ 0.005 ($U=304$, one-sided $p=0.003$, $r=0.52$, $n=20+20$).
The difference is concentrated in a minority of algorithms: four center-trained algorithms lose 0.30--0.36 and three
0.09--0.17, whereas no shift-trained algorithm loses more than 0.05. Neither the unshifted (0.612 vs.\ 0.559, two-sided
$p=0.29$) nor the shifted test scores (0.520 vs.\ 0.554, $p=0.25$) differ significantly between the groups, so we do not
claim that one group performs better or equally well on shifted functions. In both groups the training score exceeds the
independent test score by about 0.03 (0.029 in both), the expected optimism of picking the best of 200 candidates on the
training runs.

\begin{figure}[t]
\centering
\includegraphics[width=\columnwidth]{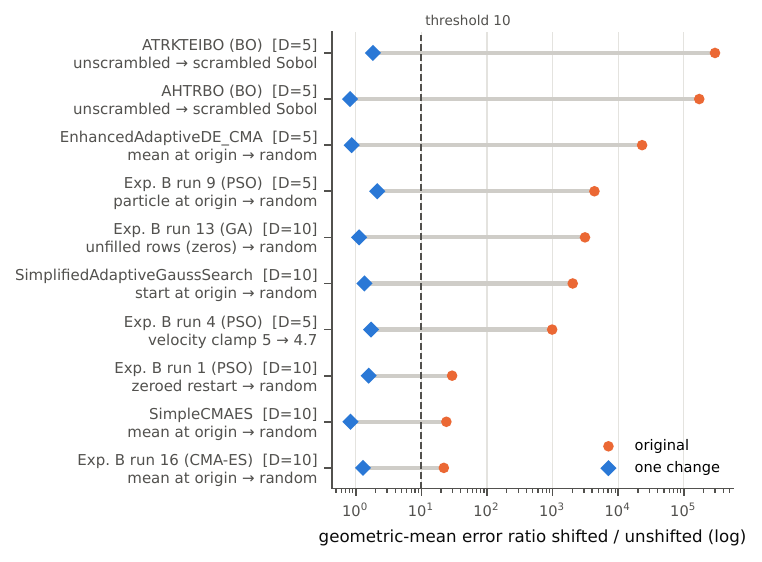}
\caption{Single-change ablations: shift ratio of the original algorithm and of the version with one change
(dimension at which the original is most biased).}
\label{fig:ablation}
\end{figure}

\begin{figure*}[t]
\centering
\includegraphics[width=0.85\textwidth]{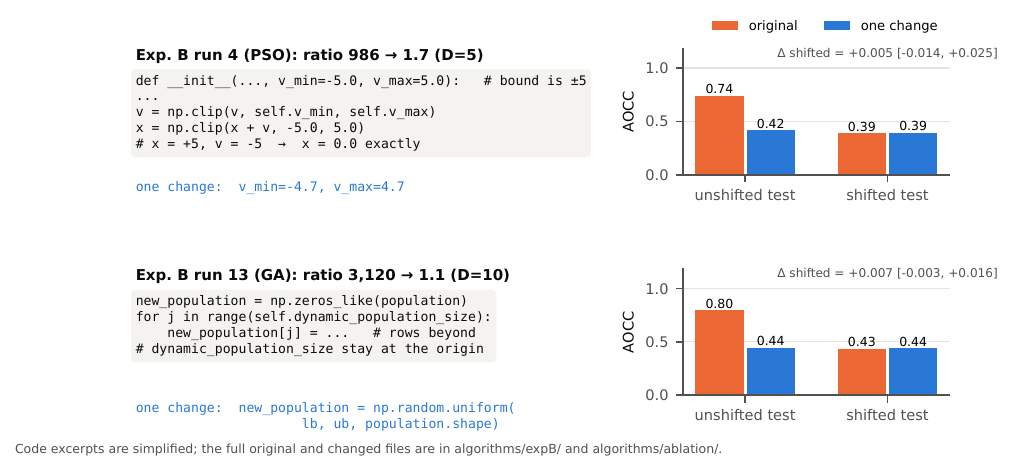}
\caption{Two cases from the centered group (code simplified). Left: the responsible code and the single change.
Right: AOCC before and after the change on the same independent test instances; $\Delta$: paired mean difference on the
shifted instances with 95\% bootstrap interval (60 paired runs).}
\label{fig:case}
\end{figure*}

\textbf{Timeline: appearance versus retention.} We screened all 497 best-so-far algorithms (Fig.~\ref{fig:timeline}).
A biased best-so-far algorithm appeared in 6 of 20 centered and 3 of 20 shifted runs ($p=0.45$, no difference detected):
biased code reached the top of the population even when the optimum was shifted, so it is not produced only by centered
training, consistent with a prior of the LLM. Because these are best-so-far algorithms, i.e.\ already selected by score,
and later proposals depend on earlier code and feedback, this is not a direct measurement of the LLM's prior.
What differed was retention. In the shifted group all three were replaced by unflagged algorithms within a few generations;
in the centered group five of six were kept to the end (5/6 vs.\ 0/3, $p=0.048$, OR 25.7; based on few events).
When a biased algorithm first became the best in a centered run, the training score jumped by 0.07--0.41 (median 0.18);
compared run by run, this jump exceeded the run's median improvement in 5 of 6 runs but was the run's largest improvement in
only 2 of 6 (Wilcoxon signed-rank on the six runs, one-sided $p=0.078$), and one shifted run (run~14) shows a similar jump.
We therefore read the timeline as follows: biased code is proposed under both training sets and is retained when the
optimum is centered. Separating the LLM's prior from selection cleanly requires scoring one candidate pool under both
benchmarks, which we leave to future work.

% =====================================================================
\section{Mechanism: One Change Removes the Flag}
\label{sec:mechanism}

We took the ten biased algorithms with a plausible mechanism (the five biased final algorithms of the centered group and
five from the published library) and changed one place in the code at a time, keeping everything else identical,
then repeated the shift-ratio test with the same seeds (all variants: Supplementary Material~S5). For every algorithm one change brings the ratio from 22--299{,}202
down to 0.8--2.1 (Fig.~\ref{fig:ablation}). Four mechanisms are confirmed:
\begin{enumerate}
\item \textbf{Starting at the origin}: a zero vector as CMA mean, starting point, trust-region center or as one particle
  (six algorithms; for the BO algorithm ATRKTEIBO, an origin trust-region center and an unscrambled Sobol design each
  remove the flag on their own, and Fig.~\ref{fig:ablation} shows the Sobol change);
\item \textbf{Unscrambled Sobol sequences}, whose second point is the box center (two BO algorithms);
\item \textbf{A velocity clamp equal to the half-width of the box}: a particle on the bound with a saturated velocity
  lands \emph{exactly} on the center. Changing the clamp from 5 to 4.7 removes the bias (986 to 1.7);
\item \textbf{Bugs that put individuals at the origin}: population rows created with \texttt{np.zeros\_like} and never
  filled (3{,}120 to 1.1), and a restart that copies a just-reset global best, the zero vector, into the swarm (29.4 to 1.6).
\end{enumerate}
The ablations also rejected two of our hypotheses: in one PSO, a near-origin initialization and the velocity clamp
do not matter on their own, and the single particle placed at the origin does; in another, the
perturbation proportional to $|x|$ is not the cause.
To our knowledge, mechanisms 3 and 4 have not been reported for LLM-generated optimizers, and none of the four is the
multiplicative contraction typical of biased human-designed metaheuristics such as GWO.

\begin{table}[t]
\centering
\caption{Ablations of the center-trained algorithms on the independent test instances ($D=5$, AOCC). Ratio: shift ratio
at the dimension where the original is most biased. Gap: unshifted minus shifted AOCC. $\Delta$: paired change of the
shifted AOCC with 95\% bootstrap interval. Run~6 is not flagged.}
\label{tab:ablation}
\setlength{\tabcolsep}{3pt}
\begin{tabular}{llrrr}
\toprule
Run & Change & Ratio & Gap & $\Delta$ shifted\\
\midrule
13 & fill zero rows & 3120$\to$1.1 & .36$\to$.00 & $+.007$ $[-.003,+.016]$\\
4 & clamp $5\to4.7$ & 986$\to$1.7 & .35$\to$.02 & $+.005$ $[-.014,+.025]$\\
4 & clamp $5\to1$ & 986$\to$1.1 & .35$\to$.02 & $+.120$ $[+.091,+.150]$\\
9 & origin particle & 4347$\to$2.1 & .34$\to$.19 & $-.023$ $[-.068,+.018]$\\
9 & all three lines & 4347$\to$1.1 & .34$\to$.05 & $+.020$ $[-.023,+.064]$\\
1 & reset best & 29.4$\to$1.6 & .09$\to$.04 & $-.014$ $[-.031,+.001]$\\
16 & random CMA mean & 22.0$\to$1.3 & .17$\to$.07 & $-.265$ $[-.352,-.178]$\\
6 & clamp $5\to4.7$ & 6.6$\to$2.1 & .30$\to$.02 & $+.012$ $[-.008,+.033]$\\
\bottomrule
\end{tabular}
\end{table}

\textbf{Does the shortcut have real value?} We compared each original algorithm with its changed versions on identical
test instances (Table~\ref{tab:ablation}, Fig.~\ref{fig:case}). For three of the five center-trained algorithms the change
costs no detectable performance on shifted instances: the paired difference is $+0.007$ $[-0.003,+0.016]$ (run~13),
$+0.005$ $[-0.014,+0.025]$ (run~4) and $+0.020$ $[-0.023,+0.064]$ (run~9; this is the version with all three suspect lines
changed, not the single change that removes the flag); for run~1 it is a
small loss at the edge of detectability, $-0.014$ $[-0.031,+0.001]$. Intervals are 95\% bootstrap intervals over the 60
paired (function, run) instances; resampling runs within each function instead gives the same conclusions. Meanwhile the gap between unshifted and shifted instances shrinks by 0.29--0.36 (0.04 for run~1).
The same holds for an unflagged algorithm with the velocity-clamp mechanism (run~6). For these algorithms the biased code
raised the score only where the optimum is at or near the center. Other changes do cost performance: a CMA-ES started at
the center of the box drops from 0.50 to 0.23 on shifted instances when started at a random point ($-0.265$
$[-0.352,-0.178]$). Starting at the center is a sensible design when optima are spread over the box, so the relevant
criterion is not whether code refers to the center but whether it still helps when the optimum moves. Some changes even help:
reducing run~4's velocity clamp to 1 raises its shifted score by 0.12 $[0.09,0.15]$.
Shift ratio and AOCC also measure different aspects: removing only the origin particle in run~9 removes the flag on final
error (ratio 2.1) but leaves an AOCC gap of 0.19 (from 0.34) and changes the shifted score by $-0.023$ $[-0.068,+0.018]$, because its initialization concentrated near the origin still speeds up early
convergence on unshifted functions; passing a final-error threshold does not remove all position dependence of the search.

% =====================================================================
\section{Discussion}
\label{sec:discussion}
\textbf{Why is bias rare in the published LLM-generated algorithms we collected?} It is not rare in the metaheuristics the
LLMs learned from: nearly half of two widely used libraries are biased under the same test. The large benchmark suites used by LLM-AAD
(BBOB~\cite{hansen2021coco}, MA-BBOB~\cite{vermetten2023mabbob}, SBOX-COST) mostly place the optimum away from the center, so a center shortcut would rarely pay off; this is a plausible explanation, consistent with our controlled experiment, in which the same LLM yielded biased survivors only when the training optimum was centered.
Why the few published exceptions arose we cannot tell from our data; for the two BO algorithms, the center point
produced early by an unscrambled Sobol design may simply be a good first sample on average when optima are spread over the box,
which would make it useful rather than exploitative (cf.\ the CMA-ES case in Section~\ref{sec:mechanism}).

\textbf{Recommendations.} (1) Train with randomly shifted optima; in our experiment, with shifts re-drawn for every run, no final algorithm trained
this way was flagged. (2) Report a shift-ratio or held-out shifted score
for every proposed algorithm; BIAS alone conflates boundary and center bias. (3) Check generated code for starting
points at the origin, unscrambled Sobol designs, and velocity clamps equal to the box half-width. (4) Do not exclude
instances on which an algorithm ``cheats''; report them.

\textbf{Limitations.} The human-designed library consists of two Python libraries with default parameters; their
implementations can differ from the original publications, and the 136 base algorithms of mealpy appear with several variants
each. In the shifted group a new shift is drawn for every run of every candidate, so the intervention changes both the location
of the optimum and the variability of the training instances; a group with one shift per evolution run, shared by all its
candidates, would separate the two. The controlled experiment uses
one LLM and 20 runs per group, and its sample size was extended after an interim look; the Fisher comparison of final
algorithms is not significant after correction, and the timeline rests on 6 and 3 events of already-selected best-so-far
algorithms, so it does not measure the LLM's prior directly. The independent test uses unseen instances of the training
functions, not new function families, and five runs per function. Qwen2.5-Coder hard-codes the box
$[-5,5]$, so the asymmetric box cannot separate origin from center bias for these algorithms. Intermediate algorithms are
screened with 10 runs, so verdicts near the threshold are uncertain. The LLM was not run with a fixed sampling seed; all
calls are logged instead.

% =====================================================================
\section{Conclusion}
In the published LLM-generated optimizers we collected, the shift-ratio test (at $D=5$ and $10$, with K\r{u}dela's
threshold) rarely detects a center or origin bias,
whereas nearly half of 330 human-designed metaheuristics are biased under the same protocol. Our controlled experiment suggests that this is a property of the training benchmarks rather than an immunity of LLM-AAD. Biased code reaches the top of the population under both training sets, but a benchmark with the optimum at the center keeps it, and such algorithms lose part of their apparent quality when the optimum moves. In each of the ten biased algorithms we examined, changing one code location removed the flag; in two of them the
responsible code is a bug. In our controlled experiment, randomly shifting the training functions kept flagged algorithms from
surviving to the final generation.

\balance
\bibliographystyle{IEEEtran}
\bibliography{refs}

% =====================================================================
\clearpage
\onecolumn
\raggedbottom
\begin{center}
{\LARGE Supplementary Material}\\[4pt]
{\large The LLM Proposes, the Benchmark Disposes: Center Bias in LLM-Driven Algorithm Design}
\end{center}
\vspace{6pt}
\setcounter{table}{0}\renewcommand{\thetable}{S\arabic{table}}
\setcounter{figure}{0}\renewcommand{\thefigure}{S\arabic{figure}}

\section*{S1. Deviations from K\r{u}dela's protocol}
K\r{u}dela~\cite{kudela2023noone} uses $D=30$, $5\times10^4$ evaluations, 20 runs and each function's native box.
We use $D\in\{5,10\}$, $2000D$ evaluations and 30 runs, and express every function in coordinates rescaled to $[-5,5]^D$,
because the LLM-generated algorithms were designed for that box and many hard-code it. The 10\% shift, the error floor of
$10^{-8}$ and the threshold of 10 on the geometric mean are unchanged. The asymmetric-box version is our addition.
Runs that fail (exception, time limit, or no in-bounds evaluation) are excluded from the mean error; a verdict is reported as
invalid if any function has fewer than 10 successful runs in either version. Recomputing every valid verdict with only the
seeds that succeeded in both versions, or with failed runs set to the worst error of that function, changes none of them.
Runs were distributed over several machines (x86 Linux, Apple silicon, Windows). Re-running stored tasks on a different
machine reproduces them exactly on the same CPU type; across CPU types the best value found is identical in our checks, while
the best point can differ in the last digits where several points share the best value, and libraries can differ in the
eighth significant digit; no verdict depends on such differences.

\section*{S2. Sources of the LLM-generated library}
\begin{table}[H]
\centering
\caption{Sources of the LLM-generated algorithms (Section~\ref{sec:methods}). ``Run best'': the best algorithm of a
published LLM-AAD run; ``named'': an algorithm presented in the publication.}
\begin{tabular}{lrrl}
\toprule
Source & Algorithms & Distinct & Selection \\
\midrule
BLADE, SBOX-COST runs~\cite{vanstein2025blade} & 199 & 199 & run best \\
BLADE, MA-BBOB runs~\cite{vanstein2025blade} & 50 & 50 & run best \\
LLaMEA paper runs~\cite{vanstein2025llamea} & 45 & 45 & run best \\
LLaMEA paper, EoH baseline~\cite{vanstein2025llamea,liu2024eoh} & 5 & 5 & run best \\
LLaMEA-BO~\cite{llameabo2025} & 20 & 17 & 5 named, 15 run best \\
LLaMEA-HPO & 8 & 8 & 3 named, 5 run best \\
LLaMEA repository & 2 & 2 & named \\
GECCO 2025/2026 LLM-designed-EA competitions & 4 & 4 & named entries; 3 screened$^\ast$ \\
\midrule
Total & 333 & 330 & 329 screened \\
\bottomrule
\end{tabular}
\end{table}
Generating LLMs: Gemini 2.0 Flash (100), Gemini 1.5 Flash (50), Qwen2.5-Coder 14B (50), Codestral (49), GPT-4o (23),
GPT-3.5 Turbo (15), GPT-4 Turbo (15), and others reported only in the source papers. $^\ast$One competition entry (MSHO-LLM) is
not screened because it calls an LLM at run time.

\section*{S3. Biased algorithms in the published LLM-generated library}
\begin{table}[H]
\centering
\caption{All LLM-generated algorithms with a geometric-mean shift ratio above 10 (standard protocol, and the BO protocol at
$D=5$). ``Origin or center'': the algorithm ignores the given bounds, so the asymmetric box cannot separate the two.
Mechanisms marked $\dagger$ are confirmed by the ablations of Section~\ref{sec:mechanism} and Table~\ref{tab:s-ablation}.}
\small
\resizebox{\textwidth}{!}{%
\begin{tabular}{llrrll}
\toprule
Algorithm & LLM & $D=5$ & $D=10$ & Type & Mechanism \\
\midrule
ATRKTEIBO\_VAD\_SuccessAware (BO) & see~\cite{llameabo2025} & 299{,}202 & -- & origin or center & zero trust-region center + unscrambled Sobol$^\dagger$ \\
AHTRBO\_EI\_DBS (BO) & see~\cite{llameabo2025} & 172{,}112 & -- & origin or center & unscrambled Sobol (second point = center)$^\dagger$ \\
EnhancedAdaptiveDE\_CMA & Gemini 2.0 Flash & 23{,}151 & 14{,}736 & origin & CMA mean starts at the origin$^\dagger$ \\
SimplifiedAdaptiveGaussSearch & Codestral & 450 & 2{,}029 & origin or center & starting point at the origin$^\dagger$ \\
CoordinateCMAES & Gemini 2.0 Flash & 48.7 & 4.6 & center & not identified \\
SimpleCMAES & Gemini 2.0 Flash & 12.0 & 24.1 & origin & mean starts at the origin$^\dagger$ \\
EnhancedFireworkAlgorithmWithLocalSearch & GPT-3.5 Turbo & 11.5 & 1.3 & origin or center & not identified \\
AdaptiveNelderMead & Gemini 1.5 Flash & 10.3 & 1.2 & center & not identified \\
MultiStageHybridGradientBoostedAnnealing & GPT-4o & 9.6 & 22.2 & origin & not identified \\
EnhancedCooperativePSO & Gemini 2.0 Flash & 1.05 & 11.6 & center & not identified \\
\bottomrule
\end{tabular}}
\end{table}

\section*{S4. Settings of the controlled evolution (experiment B)}
\begin{itemize}
\item LLaMEA 1.4.2, default $(5{+}5)$ strategy, 200 LLM-generated candidates per run, 20 independent runs per group;
  run seed $4242+100\cdot\textit{run}+\textit{group}$.
\item LLM: Qwen2.5-Coder-14B~\cite{hui2024qwen25coder} served locally by Ollama, temperature 0.8; every call (prompt,
  response, token counts) is logged.
\item Training set: F01--F13 without F08, $D=5$, $[-5,5]^5$, budget $10^4$ evaluations, two runs per function.
  Score: mean over functions of the AOCC of the best-so-far in-bounds $\log_{10}$ error, clipped to $[10^{-8}, e_{\max}]$,
  where $e_{\max}$ is the median error of 200 uniformly random points.
\item Centered group: functions unshifted, optimum at the origin (the center of the box) for 8 of the 12 functions and within
  0.17 of it per coordinate for the other four (Rosenbrock $1/6$, Step $-0.025$, Penalized~1 $-0.1$, Penalized~2 $0.1$, in
  $[-5,5]$ coordinates). Shifted group: a new random shift for every optimization run of every candidate, fixed during the
  run, optimum (also for the four functions whose unshifted optimum is not at the origin) uniform in $[-4,4]^D$.
\item Each candidate is evaluated in a separate process with a per-run limit of 60\,s and a hard limit of 1300\,s; five runs
  that crashed because of a time-out in the framework were re-run after the fix (crash logs retained).
\item Independent test: every final algorithm and every ablation variant runs on the \emph{same} pre-generated instances:
  the 12 training functions unshifted and with one random shift per function and run (shift seed fixed per function and run),
  5 runs each, with optimizer seeds 7000+ and noise seeds 9000+ (training used 1000+ and 1017+). These are unseen instances
  of the training functions, not new functions. An earlier version of this test reused the training seeds for the unshifted
  case, which reproduced the centered group's training scores exactly, and derived the shifts from the algorithm's name, so
  ablation variants were compared on different instances; both flaws are corrected here (raw runs:
  \texttt{results/heldout\_v2\_runs.csv}).
\end{itemize}
Task prompt (identical for both groups):
\begin{lstlisting}
The optimization algorithm should handle a wide range of tasks, which is evaluated on a suite of 12 continuous benchmark functions. Your task is to write the optimization algorithm in Python code to minimize the function value. The code should contain an `__init__(self, budget, dim)` function and the function `def __call__(self, func)`, which should optimize the black box function `func` using `self.budget` function evaluations.
The func() can only be called as many times as the budget allows, not more. Each of the optimization functions has a search space between -5.0 (lower bound) and 5.0 (upper bound). The dimensionality can be varied.

Give an excellent and novel heuristic algorithm to solve this task.
\end{lstlisting}

\section*{S5. All ablation variants}
\begin{table}[H]
\centering
\caption{Every variant and its shift ratio (original $\rightarrow$ changed), with the same seeds as the original.
``All'': every change listed for that algorithm at once. BO algorithms are tested at $D=5$ only.}
\label{tab:s-ablation}
\small
\resizebox{\textwidth}{!}{%
\begin{tabular}{llrrl}
\toprule
Algorithm & Change & $D=5$ & $D=10$ & Conclusion \\
\midrule
Exp.\ B centered run 9 (PSO) & one particle at the origin $\rightarrow$ random & 4{,}347 $\rightarrow$ 2.1 & 3{,}673 $\rightarrow$ 1.6 & cause \\
 & initialization concentrated near the origin $\rightarrow$ uniform & 4{,}347 $\rightarrow$ 6{,}692 & 3{,}673 $\rightarrow$ 4{,}476 & not the cause \\
 & velocity clamp 5 $\rightarrow$ 4.7 & 4{,}347 $\rightarrow$ 2{,}687 & 3{,}673 $\rightarrow$ 3{,}124 & not the cause \\
 & all three & 4{,}347 $\rightarrow$ 1.1 & 3{,}673 $\rightarrow$ 1.2 & \\
Exp.\ B centered run 13 (GA) & unfilled population rows (zeros) $\rightarrow$ random & 1{,}797 $\rightarrow$ 0.9 & 3{,}120 $\rightarrow$ 1.1 & cause (bug) \\
Exp.\ B centered run 4 (PSO) & velocity clamp $\pm5$ $\rightarrow$ $\pm4.7$ & 986 $\rightarrow$ 1.7 & 149 $\rightarrow$ 1.8 & cause \\
 & velocity clamp $\pm5$ $\rightarrow$ $\pm1$ & 986 $\rightarrow$ 1.1 & 149 $\rightarrow$ 1.1 & \\
Exp.\ B centered run 16 (CMA-ES) & mean starts at the origin $\rightarrow$ random & 7.0 $\rightarrow$ 1.7 & 22.0 $\rightarrow$ 1.3 & cause \\
Exp.\ B centered run 1 (PSO) & global best reset to the origin $\rightarrow$ random & 1.6 $\rightarrow$ 6.1 & 29.4 $\rightarrow$ 1.6 & cause (bug) \\
 & perturbation $\propto |x|$ $\rightarrow$ constant & 1.6 $\rightarrow$ 35.1 & 29.4 $\rightarrow$ 34.7 & hypothesis rejected \\
Exp.\ B centered run 6 (PSO, unflagged) & velocity clamp 5 $\rightarrow$ 4.7 & 4.8 $\rightarrow$ 0.3 & 6.6 $\rightarrow$ 2.1 & same mechanism as run 4 \\
EnhancedAdaptiveDE\_CMA & CMA mean at the origin $\rightarrow$ random & 23{,}151 $\rightarrow$ 0.9 & 14{,}736 $\rightarrow$ 1.1 & cause \\
SimplifiedAdaptiveGaussSearch & starting point at the origin $\rightarrow$ random & 450 $\rightarrow$ 0.8 & 2{,}029 $\rightarrow$ 1.4 & cause \\
SimpleCMAES & mean at the origin $\rightarrow$ random & 12.0 $\rightarrow$ 1.1 & 24.1 $\rightarrow$ 0.8 & cause \\
ATRKTEIBO (BO) & trust-region center at the origin $\rightarrow$ random & 299{,}202 $\rightarrow$ 0.9 & -- & cause \\
 & unscrambled $\rightarrow$ scrambled Sobol & 299{,}202 $\rightarrow$ 1.8 & -- & cause \\
 & both & 299{,}202 $\rightarrow$ 1.1 & -- & \\
AHTRBO (BO) & unscrambled $\rightarrow$ scrambled Sobol & 172{,}112 $\rightarrow$ 0.8 & -- & cause \\
\bottomrule
\end{tabular}}
\end{table}

\begin{table}[H]
\centering
\caption{AOCC of the experiment-B algorithms before and after each change on the same independent test instances
(12 training functions $\times$ 5 runs, unshifted and shifted; Supplementary Material~S4). $\Delta$: paired mean difference
(changed $-$ original) over the 60 paired runs, with 95\% bootstrap interval; ``gap'' = unshifted $-$ shifted.}
\label{tab:s-heldout}
\resizebox{\textwidth}{!}{%
\begin{tabular}{llrrrr}
\toprule
Algorithm & Change & Unshifted test & Shifted test & $\Delta$ shifted [95\% CI] & $\Delta$ gap [95\% CI] \\
\midrule
run 13 & unfilled rows $\rightarrow$ random & 0.796 $\rightarrow$ 0.442 & 0.434 $\rightarrow$ 0.441 & +0.007 [$-$0.003, +0.016] & $-$0.360 [$-$0.434, $-$0.286] \\
run 4 & velocity clamp 4.7 & 0.738 $\rightarrow$ 0.416 & 0.388 $\rightarrow$ 0.394 & +0.005 [$-$0.014, +0.025] & $-$0.328 [$-$0.397, $-$0.256] \\
run 4 & velocity clamp 1 & 0.738 $\rightarrow$ 0.531 & 0.388 $\rightarrow$ 0.508 & +0.120 [+0.091, +0.150] & $-$0.327 [$-$0.406, $-$0.250] \\
run 6 (unflagged) & velocity clamp 4.7 & 0.776 $\rightarrow$ 0.511 & 0.476 $\rightarrow$ 0.488 & +0.012 [$-$0.008, +0.033] & $-$0.277 [$-$0.354, $-$0.202] \\
run 1 & global best reset $\rightarrow$ random & 0.674 $\rightarrow$ 0.616 & 0.587 $\rightarrow$ 0.573 & $-$0.014 [$-$0.031, +0.001] & $-$0.044 [$-$0.089, $-$0.003] \\
run 1 & perturbation $\propto|x|$ $\rightarrow$ constant & 0.674 $\rightarrow$ 0.672 & 0.587 $\rightarrow$ 0.584 & $-$0.003 [$-$0.019, +0.014] & +0.001 [$-$0.020, +0.022] \\
run 9 & origin particle $\rightarrow$ random & 0.911 $\rightarrow$ 0.737 & 0.575 $\rightarrow$ 0.552 & $-$0.023 [$-$0.068, +0.018] & $-$0.151 [$-$0.232, $-$0.073] \\
run 9 & near-origin init.\ $\rightarrow$ uniform & 0.911 $\rightarrow$ 0.917 & 0.575 $\rightarrow$ 0.575 & $-$0.000 [$-$0.048, +0.047] & +0.006 [$-$0.040, +0.055] \\
run 9 & velocity clamp 4.7 & 0.911 $\rightarrow$ 0.907 & 0.575 $\rightarrow$ 0.553 & $-$0.022 [$-$0.057, +0.012] & +0.018 [$-$0.016, +0.051] \\
run 9 & all three changes & 0.911 $\rightarrow$ 0.640 & 0.575 $\rightarrow$ 0.595 & +0.020 [$-$0.023, +0.064] & $-$0.291 [$-$0.384, $-$0.203] \\
run 16 & mean $\rightarrow$ random start & 0.668 $\rightarrow$ 0.300 & 0.496 $\rightarrow$ 0.231 & $-$0.265 [$-$0.352, $-$0.178] & $-$0.104 [$-$0.198, $-$0.011] \\
\bottomrule
\end{tabular}}
\end{table}

\end{document}